# Predicting environment effects on breast cancer by implementing machine learning

Muhammad Shoaib Farooq, Mehreen Ilyas
Department of Computer Science, School of System and Technology, University of Management and Technology, Lahore, 54000
Corresponding author: Muhammad Shoaib Farooq (Shoaib.farooq@umt.edu.pk)

**ABSTRACT** The biggest Breast cancer is increasingly a major factor in female fatalities, overtaking heart disease. While genetic factors are important in the growth of breast cancer, new research indicates that environmental factors also play a substantial role in its occurrence and progression. The literature on the various environmental factors that may affect breast cancer risk, incidence, and outcomes is thoroughly reviewed in this study report. The study starts by looking at how lifestyle decisions, such as eating habits, exercise routines, and alcohol consumption, may affect hormonal imbalances and inflammation, two important factors driving the development of breast cancer. Additionally, it explores the part played by environmental contaminants such pesticides, endocrine-disrupting chemicals (EDCs), and industrial emissions, all of which have been linked to a higher risk of developing breast cancer due to their interference with hormone signaling and DNA damage. Algorithms for machine learning are used to express predictions. Logistic Regression, Random Forest, KNN Algorithm, SVC and extra tree classifier. Metrics including the confusion matrix correlation coefficient, F1-score, Precision, Recall, and ROC curve were used to evaluate the models. The best accuracy among all the classifiers is Random Forest with 0.91% accuracy and ROC curve 0.901% of Logistic Regression. The accuracy of the multiple algorithms for machine learning utilized in this research was good, which is important and indicates that these techniques could serve as replacement forecasting techniques in breast cancer survival analysis, notably in the Asia region.

## I. INTRODUCTION

The irregular cell development known as a tumor frequently spreads to other body parts. Knowing how and at what stage cancer formed in this patient is important since there are many different varieties of cancer, each of which is divided into numerous classes and categories [1]. Humans are now more susceptible than ever to developing several types of cancer. One out of every six fatalities is regarded as due to cancer, which is a primary reason for death globally. The most widespread cancer is breast cancer in terms of new cases. Over 40,920 women died in 2018 from breast cancer alone. The World Health Organization (WHO) estimates that 2.90 million women worldwide receive a breast cancer diagnosis each year. More than 100 diseases that affect various parts of the human body are referred to as cancer [2]. While it is well-known that genetic variables can increase a woman's risk of developing breast cancer, recent studies have shown how important environmental factors are to the progression of the condition. Numerous environmental factors, such as exposure to air pollution, endocrine-disrupting chemicals, socioeconomic status, and geographic location, affect the chance of developing breast cancer. In the early stages of cancer, there are a variety of unique approaches that scientists have discovered to predict the efficacy of therapy. The advancement of medicine and healthcare technology has led to a plethora of data about this matter. Here, we propose a machine-learning approach centered round patient data that was previously gathered from a large number of patients. In recent years, there has been a surge when applying machine learning techniques in the healthcare area as a means of making accurate diagnoses and classifying patients' conditions. More recently, Strategies for machine learning have been employed in healthcare to better aid in the detection and treatment of cancer. Tumors may now often be diagnosed by screenings and diagnostics before the patient even realizes anything is wrong.

This section looks at the particular environmental elements that machine learning algorithms take into account, the difficulties in gathering correct environmental data, and techniques to assess how much each one contributes to the prediction of breast cancer risk [11,12]. Machine learning algorithms have recently shown outstanding ability in evaluating big datasets and spotting complex patterns. Early detection and tailored treatment could be transformed by using these effective tools to forecast breast cancer risk based on changes in environmental factors. The integration of machine learning algorithms with environmental data to predict breast cancer risk is examined in this study, offering insight on how these cutting-edge methods can help identify high-risk populations and enhance preventive measures [9,10].Machine learning algorithms have the potential to completely change the way we provide care for patients with breast cancer. These cutting-edge computational technique may examine enormous datasets that cover a wide range of environmental characteristics, such as air quality, socioeconomic status, accessibility to healthcare services, and more. We can find hidden patterns, correlations, and risk factors by using these algorithms to analyze this data that may have eluded detection using only conventional statistical techniques. This newly discovered knowledge of how the environment influences breast cancer survival improves therapeutic decision-making and provides hope for focused interventions and individualized treatment regimens, ultimately benefiting the lives of numerous breast cancer patients. Predictions are generated using learning-based classification models, and their performance is measured against test data. This conclusion was reached by comparing and analyzing several methods of categorization. Random forest algorithms are a great tool for machine learning and accuracy because of their flexibility and simplicity of usage. When several classifiers are swiftly merged in an ensemble analysis a huge collection is irrelevant. In this study, we suggest a patient-specific utilization of machine learning to increase the survival rate for breast cancer patients. The accuracy study of precision, sensitivity, and specificity, and the performance of the suggested algorithms was assessed using AUC. . Here, five machine learning techniques— including the support vector classifier (SVC), the KNN algorithm, the extra tree (ET) classifier, the logistic regression (LR), and the Random Forest—were created for identifying patient will be survived or not. The top five algorithms that have been used to identify cancer using medical datasets were passed over because of their superior performance.

The novelty of using machine learning algorithms to analyze the influence of the environment on breast cancer patient survival resides in its potential to revolutionize clinical and academic healthcare research. Traditional epidemiological studies frequently have difficulty capturing the complex and nuanced interactions between environmental factors and serious diseases like breast cancer. Machine learning, on the other hand, enables the integration of many variables, even those with non-linear and nuanced interactions, providing a more thorough viewpoint. By examining these intricate relationships, we can discover previously unknown risk factors, protective components that are unique to particular patients or subpopulations, and environmental factors that have an impact on the patient as an individual. This advances the field of breast cancer research and improves patient outcomes in a ground-breaking way by improving not only our awareness of the complex dynamics at play but also opening the door for tailored interventions, early detection techniques, and more efficient treatment regimens.

The format of our article is structured as follows: we discussed the related work in section 2.We briefly discussed the material and method in this article in section 3.Summary of Results from Experiment in section 4 is Presented. The briefly discussion in section 5. In section 6 the conclusion of our work is given. The paper ends with a list of references.

## 2. RELATED WORK

Mohammad Nazmul Haque et al.[4] The SEER Program of the National Cancer Institute, which information about cancer in the general population, was updated in November 2017 and used to build this patient record for breast cancer. 4024 patients were left after excluding those with unknown tumor size, examined regional LNs, positive regional LNs, and those with overall survival of less than one month. were excluded Eight popular machine-learning techniques for forecasting mortality rates from breast cancer recurrence were evaluated using a database. Decision tree, Random forest, the K-nearest neighbor predictive modeling, assistive vector technology, gradient boosting classifier, AdaBoost classifier, and voting classifier. Classification is produced by a classifier using a random forest. One of every algorithm, it was the most accurate (94.64 percent). In this case, logistic regression had an accuracy rate of 81%. The accuracy of the support vector machine is 85%. 88% of the traits connected to breast cancer can be successfully determined via the voting classifier model. The accuracy is 88% as a result, which is superior to logistic regression and support vector machine. With an accuracy of 89%, the decision tree classifier model outperforms the LR, SVM, and voting classifier. The accuracy of the K-NN model is 84%, which is higher than that of logistic regression. The GB model has a 92 percent accuracy rate. The decision tree classifier's outcome is identical to the 89% accuracy of the AdaBoost model.

Aditi kajala at el. [5] The work that is being presented makes use of the SEER, which stands for Surveillance, Epidemiology, and End Results dataset. The collection contains 4024 records of breast cancer participants with 16 attributes, 3408 records of alive cases, and 616 records of deaths. During data cleaning, the label encoding method is employed to transform all category attributes into numerical values. The prediction made by machine learning algorithms aids in understanding how different attributes affect the prediction. It can be described as a partial plot or summary plot of Shap values in graph-like structures. By using data sampling techniques, some minority or majority samples are either expanded or removed while maintaining the necessary relevant information. Replicating samples from minority classes in order to balance the dataset is referred to as oversampling. In the interest of balancing the dataset, under-sampling includes deleting samples from the dominant class. In order to balance the dataset for this investigation, four oversampling techniques—SMOTE, ADASYN, Random oversample, and SMOTE with Near-border and AllKNN under-sampling techniques—were used. Decision Tree, Random Forest, KNN, and SVM are four machine learning methods, are compared for performance. By calculating the Precision and AUC score using the used dataset, the effectiveness of these models was assessed. The outcome demonstrated that the SVM model with a Random overs sampler outperformed all others and obtained precision and AUC scores of 1 and 0.9935, respectively.

Aqsa Rahim at el.[6] The numerous machine learning models that various writers have presented and their successes. According to the method outlined in this work, just one highly correlated feature, i.e., if three characteristics are highly linked, is chosen for feature selection. This is done by first eliminating the highly associated features. A smaller feature set of 16 features is the result of this. reducing the number of capabilities, even more, I have used the recursive feature selection method. As a result, the classification models receive the selection of the top 11 features for categorization. Random forest achieves the highest level of accuracy. The suggested approach tested the impact of the feature space through feature space selection. To accomplish this, the showcase space was condensed to 11 features using a mix of Recursive feature elimination (RFE) and connection-based picking of features. In order to address the issue of overfitting in machine learning, this is especially crucial. We contrast the results of many models that have been suggested, using techniques like SVM, Random Forest, Gradient Boosting, Artificial Neural Network, and Multilayer Perceptron Model. The Multilayer Perceptron Model, which had an accuracy of 99.12%, was the best algorithm.

Leili Tapak at el. [7] We used a dataset comprising health 550 breast cancer patient's medical records. AdaBoost, Support Vector Machine (SVM), Least-square SVM (LSSVM), Naive Bayes (NB), Random Forest (RF), Least-square SVM (LSSVM), Adabag, Logistic Regression (LR), and Linear Discriminant Analysis were utilized to forecast survival and metastasis of breast cancer. Overall accuracy, likelihood ratio, sensitivity, specificity, and specificity were used to gauge how well the approaches worked. 850 patients were still living, and 85% of them did not experience metastases. In comparison to other techniques, the SVM and LDA have a higher sensitivity (73%) for directions to ensure, with an average total specificity of all methodologies of 94%. The SVM and LDA outperform other algorithms in terms of overall accuracy (93%). The LR and LDA provided the greatest overall accuracy. (86%) for identifying metastasis, whereas The maximum specificity (98%) and sensitivity (36%) were achieved by the NB and RF, respectively.

Mogana Darshini Ganggayah at el.[8] The University Malaya Medical Centre (UMMC), Kuala Lumpur, Malaysia's Breast Cancer Registry provided the 8942 breast cancer patient records that make up a sizable hospital-based dataset. The entire dataset (referred to as "all data"), which included 8066 observations and 23 rates of survival predictors, was subjected to modeling. Six techniques were used to compare the data quality: support vector machine, decision tree, random forest, neural networks, extreme boost, and logistic regression (e1071). For the model evaluation utilizing all the techniques, then the dataset was split into a training set and a testing set. Accuracy, sensitivity, specificity, and precision were all measured for each model. Also computed for each model are the Matthew correlation coefficient and the area under the receiver operating characteristic curve (AUC). In terms of model accuracy and the calibration measure, all methods produced results that were close, with decision trees producing the lowest results (accuracy = 79.8%) and random forests producing the greatest results (accuracy = 82.7%).

## 3. MATERIAL AND METHOD
This section is divided into three parts. Priorities include taking care of feature selection, preprocessing, and patient dataset description. In the second stage, prediction models that use different ML algorithms are evaluated in order to create the model that performs the best. Lastly, once the performance of each model is assessed using the assessment criteria, the best model that can predict the variables is picked. Figure 1 depicts the organization of a number of elements for selecting the optimal model.

**Table#1: Description of dataset**

### 3.1. Dataset Description:

| Sr.No. | Attribute | Range |
|---|---|---|
| 1 | Age | 30-69 |
| 2 | Race | White, black, other |
| 3 | Marital Status | Divorced, married, single, widow |
| 4 | T stage | ,separated |
| 5 | N stage | T1,T2,T3,T4 |
| 6 | 6 stage | N1,N2,N3 |
| 7 | Differentiate | 1A,2B,3A,3B,3C |
| 8 | Grade | Well, moderately, poorly, undifferentiated |
| 9 | Stage | 1,2,3,4 grade |
| 10 | Tumor size | Regional, Distant |
| 11 | **Estrogen** | <36mm,>105mm |
| 12 | **Status** | Positive, Negative |
| 13 | **Progesterone** | Positive, negative |
| 14 | status | Total examined |
| 15 | **Regional** | Positive examined |
| 16 | nodes examined Regional nodes positive Survival months Status | Alive, dead |

The collection includes many attributes that are related to the patient's medical information. On a set of data, research, and analysis were done. The information was obtained from the Kaggle website which contained both discrete and continuous characteristics. Table 1 below provides a detailed description of the data.

### 3.2. Dataset preprocessing:
Prior to classification, the used dataset is pre-processed. Preprocessing was used to clean the data and improve the accuracy of our dataset for value prediction. Cleaning and normalizing are preprocessing chores. During data cleaning, the label encoding method is employed to transform all category attributes into numerical values. No attribute in this dataset contains any missing values. Due to the use of several measuring units, the data normalization strategy (bringing values between 0 and 1) has been utilized.
➤ 16 columns and 4024 rows make up the data set.

- The dataset is continuous, and the issue is one of classification.
- Data with categorical encoding
- Verifying values for null
- Examining outliers
- Verify the relationships between the variables.
- Data division into dependent and independent variables;
- categorization into two classes (1-alive, 2-dead
- Dataset is continuous, discrete and it is classification problem.
- Check correlation between variables.
- Dividing data into dependent and independent variables.

### 3.2.1. Classifiers:

An essential element of machine learning and data analysis is the classifier. It performs the role of an intelligent decision-maker, tasked with classifying instances or data points into specified classes or categories in accordance with their traits and attributes.

They are the most popular classification algorithms for healthcare diagnosis and could produce better results, the following automated learning techniques are used in this section to predict the survival stage of patients: KNN algorithm, logistic regression, Random forest, SVC, and extra tree classifier.

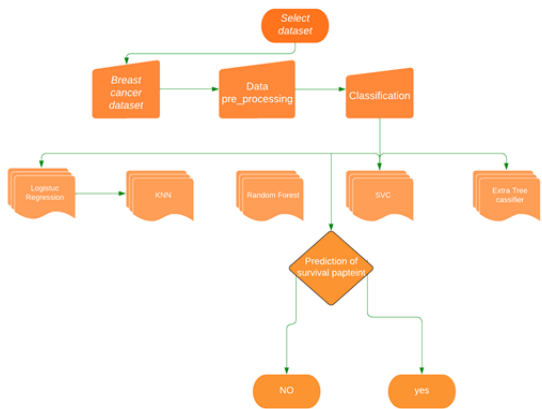

**Figure 1. Flow chart diagram of classifiers**

### 3.2.1.1. Logistic regression:

Logistic regression is a typical machine learning classification technique used to predict the probability of classes given a collection of dependent variables. All of the input characteristics are generally summed in logistic regression models to arrive at a probability estimate for the output. For a binary classification problem, the output of logistic regression is always in the range of (0, 1).

Learning, optimization, and training input data and parameters. The predicted value is produced when the input is near to 1.

$$h_{\Theta(x)} = \frac{1}{1+e^{-\theta x}}$$

If we want optimal performance on our work, we must use a loss function (also known as a cost or objective function). It is conventional to use the log-likelihood loss function in logistic regression.

m=number of samples in the training data.is given.
$y^i$ is the label of the i(th) sample.
$P^i$ such as i is the prediction value of i(th) sample. (jessica, 2022)

$$J_{(\Theta)} = -\frac{1}{m} \sum_{m}^{i=1}(y^i \log(p^i) + (1 - y^i)\log(1 - p^i))$$

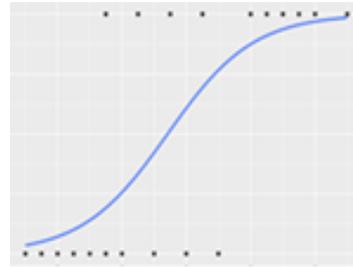

**Figure 2. Diagram of logistic regression**

### 3.2.1.2. Extra tree:

For the purpose of selecting features, the suggested methodology incorporates an embedded technique called as additional trees classifier. As a bridge when feature extraction and categorization are interrelated, it is carried out. The technique of automatically identifying the significant features that will provide the prediction variable with the most information is known as feature selection. The accuracy of the model will be reduced as a result of processing irrelevant features, which will also lengthen calculation time. The duration of the classification process in this study has been significantly decreased by the feature selection utilizing extra trees classifier [3].

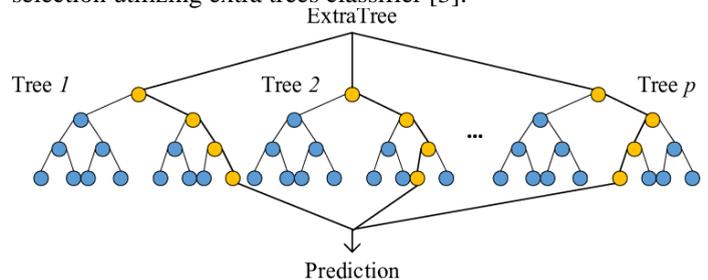

**Figure 3. Diagram of Extra Tree**

### 3.2.1.3. Random Forest:

The RF algorithm is often used while bagging. For training a classifier, traditional decision trees use the same attributes and the same subset of the data, while RF randomly picks both. Each trained classifier produces a separate set of predictions for the same input. Voting on each trained classifier's output yields the final prediction result, which is frequently done using the plurality or mean. The random feature distribution of the algorithm will increase the variety of its classifiers, improving the generalizability of the model.

To convert them to a value between 0 and 1, just divide by the total importance of all the traits.

$$normfi_i = \frac{fi_i}{\sum_{j \in allfeatures} fi_j}$$

The importance of each feature is then estimated by averaging its weight across all of the trees in the Random Forest. The relative relevance of each attribute to each tree may be calculated by adding up all the values and then dividing by the total number of trees.

$$RFfi_i = \frac{\sum_{j \in alltrees} normfi_{ij}}{T}$$

T = Number of trees overall.
$RFfi_i$ Sub (i) = the Random Forest model's estimation of the feature's importance based on data from all trees.
$normfi_{ij}$ Sub (ij) = significance of feature i for tree j normalized.

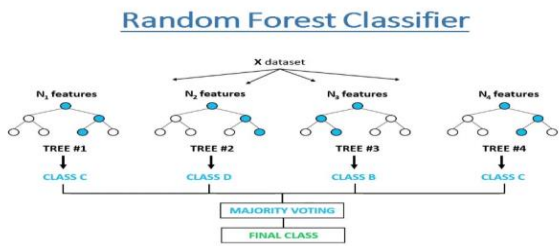

**Figure 4. Diagram of Random forest**

### 3.2.1.4. K-Nearest Neighbor:

Many of these categorization issues may potentially be solved using the supervised machine learning technique known as the k-nearest neighbor method (KNN). KNN takes the average distance between a query and each data sample to find the K most similar instances, and then uses majority voting or an average of the labels to decide on a final classification. Equation 1 represents the calculation of the distance between a Euclidean object and the provided data for training.

Euclidean distance
$$= d(x,y) = \sqrt{\sum_{i=1}^{k}(x_i - y_i)^2} \quad (1)$$

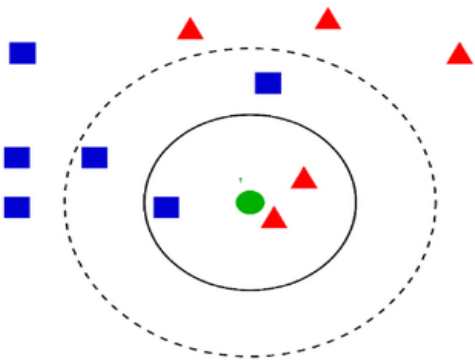

**Figure 5. Diagram of K-nearest neighbor**

### 3.2.1.5. Support vector Classifier:

The support vector classifier is one of the top supervised machine learning approaches for classification. (SVC), which has the benefit of not having the overfitting problem. In the vast majority of instances, it also gives improved categorization accuracy. Due to the binary nature of the basic SVM, multi-class classification is done by applying the "one-versus-one" method. SVC creates (n(n 1))/2 classifiers, each of which can handle data from two different classes, if there are ′n classes. With the input feature vector gathered from leukocyte pictures, the SVC creates a decision surface. [9]

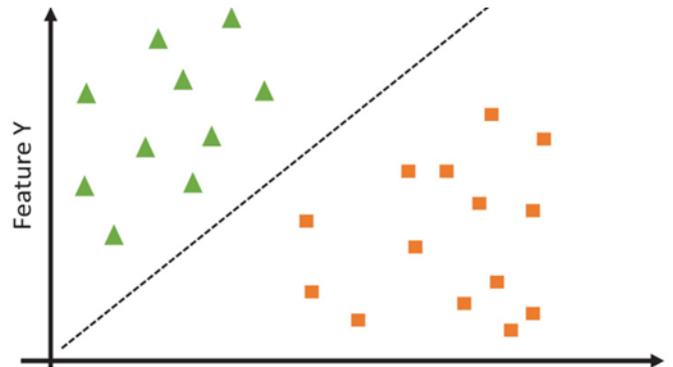

**Figure 6. Diagram of Support Vector Classifier**

### 4. Experiment and Result:

A variety of approaches are used to analyze and forecast Breast Cancer caused by Environment effects. Analytics-based strategies can provide an accurate prediction for a specific disease by grouping individuals with similar symptoms.

The study Breast cancer patient survival looked at the KNN, SVC, Logistic Regression, Extra Tree Classifier, and RF algorithms for machine learning classification. The training data from the dataset are divided by 80%, and the test data are divided by 20%.

### 4.1 Evaluation Matrix:

Accuracy, recall, and precision, f1, MCC, and ROC AUC are just a few of the helpful metrics that may be applied to both positive and negative sample predictors. The accuracy score indicates the percentage of cases and controls for which lung cancer was accurately identified. The difference between recall and precision is that the latter gives the precise number of instances of the positive class, while recall gives an estimate of how near you were to the true positive class total. For your convenience, many formulas to compute these variables are provided below.

**ACCURACY:**
The degree to which a system can accurately foretell the future is one metric by which to evaluate it. The reliability of our model is determined by how often its forecasts come true. The following formula may be used to evaluate the precision of a yes/no response.

Accuracy=$\frac{AMOUNT\ OF\ TRUE\ PREDICTIONS}{NUMBER\ OF\ PROJECTIONS\ IN\ TOTAL}$

Accuracy has the following formal definition or binary classification.

$$Accuracy=\frac{TP+TN}{TP+TN+FP+FN}$$

**PRECISION:**
Recall is also known as positive predictive value (PPV). A test's accuracy is equal to its genuine positives plus its erroneous negatives positives.

$$Precision=\frac{TP}{TP+FP}$$

**RECALL:**
Keep in mind that recall may also be also known as the true positive rate or sensitivity. The percentage of correct diagnoses to false negatives is a common metric used to evaluate diagnostic accuracy.

RECALL=$\frac{TP}{TP+FN}$

**F1_SCORE:**
The harmonic mean of precision is the F1_score.

F1_SCORE= $2*\frac{PRECISION*RECALL}{PRECISION+RECALL}$

**MCC:**
When applied to larger and larger datasets, the Matthews correlation coefficient (MCC) proves to be a more reliable statistical measure by giving more weight to the actual positives and negatives.

$$MCC=\frac{TP*TN-FP*FN}{\sqrt{(TP+FP)(TP+FN)(TN+FP)(TN+FN)}}$$

**4.2. Analysis of Result:**
This study's objective is to pinpoint the machine learning model that achieves the best results across a variety of metrics such as area under the curve, F1 Score, MCC, Precision, Recall, and Accuracy (AUC). Any model we recommend is guaranteed to be at least 80% correct.

**Correlation matrix:**
A table displaying the correlations between the various factors. A chart outlining the relationships between the various variables. The correlation between any two variables is displayed in each table cell. Correlation matrix is used to summarize the data.

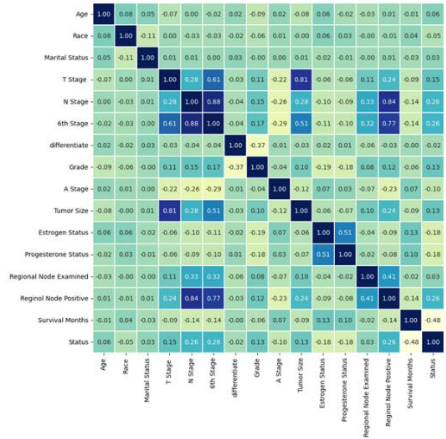

**Fig. 7 Correlation of Dataset**

The correlation coefficient has an exact range of -1 to +1. As the name suggests correlation analysis compares two variables to ascertain how closely related they are. Only correlation analysis is possible if there is a connection or likelihood between the related components. The statistical indicator of how strongly two variables are related to one another is the correlation coefficient.

Table. 2 Results of analysis or proposed methodology

| Classifiers | On confusion Matrix | | | | | Roc |
|---|---|---|---|---|---|---|
| | Acc(%) | Pre(%) | Re(%) | F1_sc(%) | MCC | Acc(%) |
| LR | 0.90 | 0.86 | 0.71 | 0.76 | 0.56 | 0.901 |
| ET | 0.89 | 0.83 | 0.70 | 0.75 | 0.53 | 0.866 |
| RF | 0.91 | 0.87 | 0.74 | 0.79 | 0.60 | 0.885 |
| KNN | 0.87 | 0.77 | 0.64 | 0.67 | 0.40 | 0.866 |
| SVC | 0.89 | 0.89 | 0.67 | 0.73 | 0.53 | 0.876 |

A comparison of the Confusion Matrix machine learning classification method is shown in Table.02. Random forest has a 0.91% accuracy. The accuracy of extra tree classifier is 0.89%, accuracy of Logistic Regression is 0.90%, accuracy of KNN classifier is 0.87% and the accuracy of support vector classifier is 0.89%. On the other side the Highest accuracy of ROC Curve is Logistic Regresion is 0.901% respectively.

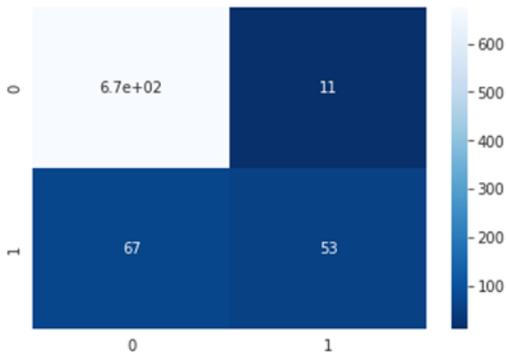

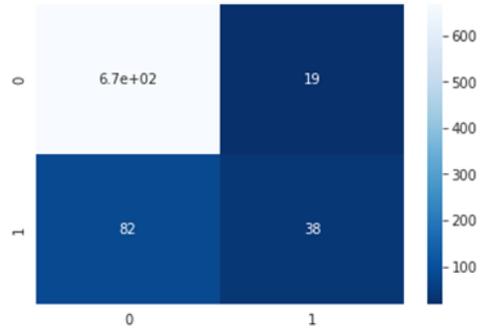

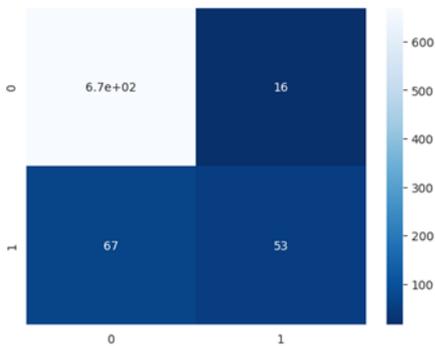

- In KNN classifier, true positive rate (TPR) is 6.7e+02% The false negative percentage (FNR) is 82%, while the false positive rate is 19% and the true negative rate (TNR) is 38%.
- In Random Forest classifier, true positive rate (TPR) is 6.7e+02% The false percentage (FNR) is 58%, and the proportion of false positives (FPR) is 13%. and the true negative rate (TNR) is 62%.

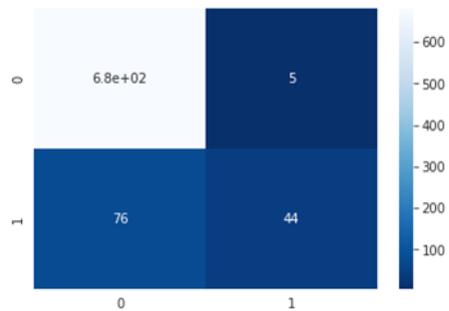

- In logistic regression true positive rate (TPR) is 6.7e+02% The false rate of negatives (FNR) is 22%, the rate of false positives (FPR) is 11% and the true negative rate (TNR) is 53%.
- In Extra tree classifier, true positive rate (TPR) is 6.7e+02% The false percentage of negatives (FNR) is 67%, and the prevalence of false positives (FPR) is 16%. and the true negative rate (TNR) is 53%.

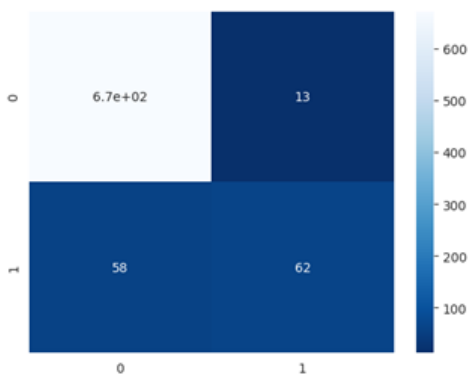

- I support Vector classifier, True Positive Rate is 6.8e+02%. The proportion of false positive is 5%. While the likelihood of false negative percentage is 76% and the true positive rate is 44%.

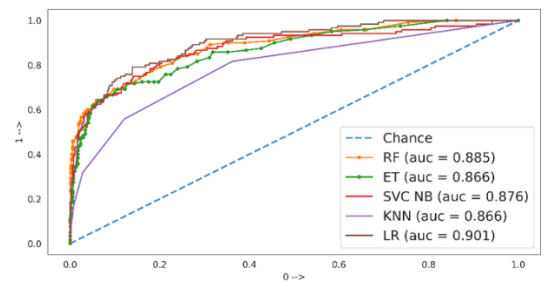

**Figure 8. Models Evaluation Based on AUC-ROC Curves**
In our study, We employed many machine-learning methods to forecast the patient survival rate for breast cancer. ROC curves are widely used to visually depict the relationship between the two when evaluating clinical sensitivity and specificity for each proposed cut-off for a test or combination of tests. The benefits of using the question test are also shown by the ROC curve's ROC area. The range of the ROC is always 0 to 1. The additional tree classifier produced results with a ROC curve of 0.901. The graph below displays the other ROC results. The area under the ROC curve reveals the use of the questioned test. The ROC always falls between 0 and 1. The Logistic Regression classifier achieves the greatest Roc curve of 0.901%. The graph depicts the other ROC results.

## 5. DISCUSSION

A cutting-edge method in research is the examination of environmental influences on breast cancer using machine learning models. A potent technique for analyzing the complex connections between the environment and breast cancer incidence, prognosis, and treatment response is machine learning. The review is predicated on the experimental work detailed below. The accuracy of a classifier may be measured first by classification. Throughout the years, several schemes for organizing data have been proposed and implemented. The "accuracy" of our trained algorithm is measured by how well it correctly guesses which categories belong to which. Classifier accuracy is compared in Table 2. Current research indicates that when used for classification challenges, Random forest outperform conventional methods (0.91% accuracy). Among the characteristics that make a random forest apart are: In many cases, random forest may achieve near-perfect accuracy. Incredible in its ability to process massive amounts of data. An approximate value is provided for a number of useful attributes that may be used to group items. This resulted in the expansion of wooded areas from which future generations may benefit. Compared to other modern technical gadgets, it lacks a flash. When there are more explanatory factors than noise variables, logistic regression is effective; otherwise, a random forest should be utilized. A random forest top-down induction model is used for classification and prediction in logistic regression, whereas path analysis is used to define the directional connections between a collection of variables. Like a group of decision trees (CARTs).

## 6. CONCLUSION AND FUTURE WORK

This paper has examined the promising combination of breast cancer data and machine learning algorithms to forecast the risk of breast cancer as the environment changes. Understanding the interactions between environmental factors and breast cancer risk is essential for enhancing prevention and early detection measures since breast cancer remains a serious worldwide health concern. Incorporating a variety of environmental parameters, including air pollution, endocrine-disrupting substances, socioeconomic situations, and geographic location, researchers can develop prediction models. Compared to previous methods, the random forest method produced somewhat higher evaluation accuracy. In this study, the accuracy rates for the RF, ET, K-NN, SVM, and LR algorithms were 0.91 percent, 0.89 percent, 0.90 percent, and 0.89 percent, respectively. All of the algorithms' accuracy appeared to be close, though. The efficiency of the model and critical factors from Environment effect on breast cancer patients' survival rates were identified in this study, which may be used in clinical practise, particularly in the context of Asia.

## REFRENCES: